\documentclass{article} 
\usepackage{iclr2023_conference_tinypaper,times}
\usepackage{graphicx}

\usepackage{amsmath,amsfonts,bm}

\def\eqref#1{equation~\ref{#1}}

\def\1{\bm{1}}

\DeclareMathAlphabet{\mathsfit}{\encodingdefault}{\sfdefault}{m}{sl}
\SetMathAlphabet{\mathsfit}{bold}{\encodingdefault}{\sfdefault}{bx}{n}

\usepackage{hyperref}
\usepackage{url}
\usepackage{subcaption}
\title{Network Inversion of Binarised Neural Nets\\}
\author{Pirzada Suhail, Supratik Chakraborty, Amit Sethi \\
IIT Bombay \\
\texttt{\{22d0518, supratik, asethi\}@iitb.ac.in} \\
}

\iclrfinalcopy 
\begin{document}
\maketitle
\begin{abstract}
While the deployment of neural networks, yielding impressive results, becomes more prevalent in various applications, their interpretability and understanding remain a critical challenge. Network inversion, a technique that aims to reconstruct the input space from the model's learned internal representations, plays a pivotal role in unraveling the black-box nature of input to output mappings in neural networks. In safety-critical scenarios, where model outputs may influence pivotal decisions, the integrity of the corresponding input space is paramount, necessitating the elimination of any extraneous "garbage" to ensure the trustworthiness of the network. Binarised Neural Networks (BNNs), characterized by binary weights and activations, offer computational efficiency and reduced memory requirements, making them suitable for resource-constrained environments. This paper introduces a novel approach to invert a trained BNN by encoding it into a CNF formula that captures the network's structure, allowing for both inference and inversion.
\end{abstract}
\section{Introduction}
The remarkable performance shown by neural networks across various applications often comes at the cost of interpretability, posing a significant challenge in their extension to safety-critical domains. The need for demystifying the internal workings of neural networks has led to the development of techniques like network inversion that unravel the black-box nature of the network by reconstructing the input space from the model's learned internal representations. The challenge of network inversion is underscored by the complex many-to-one mappings inherent in neural networks, exacerbated by the activation functions employed, making the inversion process non-trivial.

This paper proposes a novel approach to invert a Binarised Neural Network (BNN) by encoding the trained BNN into a Conjunctive Normal Form (CNF) propositional formula that comprehensively captures the network's structure, encompassing the values in the input, hidden, and output layers of the network. The CNF formula precisely encodes the computation in each neuron of the BNN, thereby making it possible to mimic the overall inference task with 100\% precision.  Interestingly, the same encoding can be used for the inversion problem simply by constraining the propositional variables corresponding to network outputs, and invoking a satisfiability (SAT) solver to find satisfying assignments for the propositional variables corresponding to network inputs.

This approach unlike other optimization-based techniques, circumvents the need for any careful hyper-parameter tuning. Additionally, the deterministic nature of CNF encodings provides fine-grained control over inverted samples. Specifically, sampling the satisfying assignments (near-)uniformly guarantees diverse input generation during inversion, addressing concerns like mode collapse commonly associated with generative models.




\section{Methodology \& Implementation}
\label{gen_inst}

The implementation follows the definition of a simple BNN adapted from \citet{courbariaux2016binarized} with a block structure wherein each block has a linear layer with binarised weights, a batch normalisation layer and a binarisation layer, ensuring that both the input to and output from the block are binarised. The output block has a linear layer followed by a softmax for classification purposes. The trained BNN is then encoded into a CNF formula using the publicly available tool NPAQ \citet{baluta2019quantitative} in which the relationship between the input, hidden \& output layers of the neural network is captured by the variables and clauses. Consider a BNN encoded into a CNF Formula $BNN(X,H,Y)$ where ${X}, {H}$ and ${Y}$ are the sets of the input, hidden and output variables. 

Inference can be performed by adding constraints on the input variables as $I(X=x)$ followed by evaluating for the satisfying assignments of \[ BNN(X,H,Y) \wedge I(X=x), \] among which the output variables would have the predicted label for the given input. Similarly, inversion can be performed by adding constraints on the output variables as $O(Y=y)$ followed by evaluating for the satisfying assignments of \[ BNN(X,H,Y) \wedge O(Y=y), \] among which the the input variables represent one of the inputs for the set label. While the satisfying assignments for the above CNF formulas can be obtained using SAT Solvers, for a more comprehensive exploration of the label's input space, CMSGen, \citet{9617576}, a uniform-like sampler, is used to sample diversely from the satisfying assignments of the constrained CNF formula.

\section{Experiments \& Results}
\label{headings}

The experimental demonstration of the inversion process is performed on a 100-20-10 architecture BNN with 100 neurons in the input for flattened 10x10 MNIST images as shown in Fig. 1, 20 in the hidden layer and 10 in the output layer, trained for 25 epochs with an accuracy of around 75\%, encoded into a CNF formula with over 60k variables and 100k clauses, wherein the variables from 1 to 100 represent the network inputs and variables from 101 to 110 represent the network outputs. The CNF formula, conjoined with an additional constraint that sets an output variable (say, variable 103 corresponding to label 2) to true, is then sampled with CMSGen for 100 satisfying assignments.

\begin{figure}[h]
\begin{minipage}{.29\textwidth}  
  \centering
  \includegraphics[width=1\linewidth]{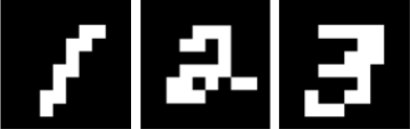}  
  \captionof{figure}{Training Images}
  \label{fig:test1}
\end{minipage}\hspace{5mm} 
\begin{minipage}{.6\textwidth}  
  \centering
  \includegraphics[width=1.11\linewidth]{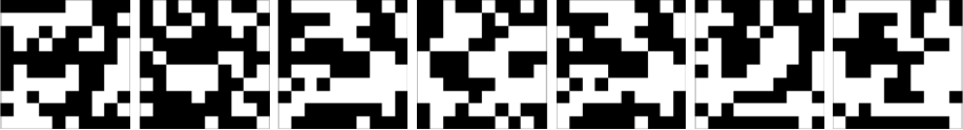}  
  \captionof{figure}{Inverted Images for Class 2}
  \label{fig:test2}
\end{minipage}
\end{figure}

Using the satisfying assignment for network input variables, we can now re-construct an input image that is guaranteed to be classified by the BNN as per the output (label) variable that was set before solving.  Example inverted images for output label "2" are shown in Fig. 2. By way of validation, all input images obtained from our network inversion process were also fed as inputs to the BNN and we checked that this yielded the same output label (e.g. label "2" for the images in Fig. 2) as expected.  As can be seen from Fig. 2, several of these input images do not resemble anything close to what the network was trained on, and what can be reasonably expected to be classified as "2".  These are erroneous classifications that can be hugely problematic in safety-critical applications, and our approach is able to unearth such classifications. In yet another experiment, an inadequately trained, 25-20-10 architecture BNN on flattened 5x5 MNIST images did not classify any image as having the label "8".  On encoding the BNN into a CNF formula with 6k variables and 13k clauses, and on trying to solve the inversion problem for label "8", we found the formula to be unsatisfiable which conclusively showed that nothing in the input space is labeled "8" by this network.

\section{Conclusion \& Future Work }
\label{others}
This paper introduces a novel approach to network inversion for BNNs by encoding them into CNF formulas. Conjoined with constraints on the output variables, the CNF formula is uniform-like sampled using CMSGen, providing a more nuanced understanding of the input space associated with specific output labels. Notably, in the inversion process, the reconstructed inputs appear unlike anything the network was trained on, highlighting their unsuitability for safety-critical applications. 

Network Inversion can be employed as an iterative tool to re-train a model using out-of-distribution inputs generated during inversion, by labeling these inputs specifically as "garbage" and by adding an extra class label for "garbage". Through successive iterations of inversion and retraining, we aim to progressively refine the model's input space, ultimately including only relevant data and mitigating the influence of irrelevant samples, thus enhancing the generalization capabilities of BNNs.

\subsubsection*{URM Statement}
The authors acknowledge that at least one key author of this work meets the URM criteria of ICLR 2024 Tiny Papers Track.
\bibliography{iclr2023_conference_tinypaper}
\bibliographystyle{iclr2023_conference_tinypaper}

\appendix
\section{Finer Implementation Details}

The finer details of the implementation of this approach includes the definition of a feed-forward BNN from \citet{courbariaux2016binarized} with a block structure. Each of the inner blocks of the BNN have in them three layers including a Binarised Linear Layer, a Batch Normalisation Layer, and a Binarisation Layer. For the \( k \)th block of the BNN, the input to and the output from each of the blocks is binarised as \( x_k \) and \( x_{k+1} \) respectively. 

The binarised linear layer defined as:
\begin{equation}
y = A_k x_k + b_k,
\end{equation}
does an affine transformation of the binarised input vector \( x_k \),  wherein the weights are restricted to either 1 or -1, while the bias is allowed to take real values. Hence, the output \( y \) is therefore a real-valued vector. 

Following the affine transformation, a batch normalization layer is applied that normalises the output \( y \) to have a mean of zero and a variance of one, across the batch of data as:
\begin{equation}
z = \frac{y - \mu_{\mathcal{B}}}{{\sigma_{\mathcal{B}} + \epsilon}} \cdot \alpha + \gamma,
\end{equation}
where \( \mu_{\mathcal{B}} \) is the batch mean, \( \sigma_{\mathcal{B}} \) is the standard deviation, \( \alpha \) is a scale parameter, \( \gamma \) is a shift parameter, and \( \epsilon \) is a small constant added for numerical stability. The parameters \( \alpha \) and \( \gamma \) are learnable and can be adjusted during the training process. 

Lastly a binarisation layer defined as: 
\begin{equation}
x_{k+1} = \text{sign}(z),
\end{equation}
maps the real-valued vector \( z \) to a binary vector \( x_{k+1} \). The sign function is applied element-wise to the vector \( z \), such that each element \( z_i \) in \( x_{k+1} \) is set to 1 if \( z_i \) is greater than or equal to 0, and -1 otherwise. 

The output block of  the BNN is composed of a similar Binarised Linear Layer and a Softmax Layer for classification purposes defined as:
\begin{equation}
o = \text{argmax}(A_n x_n + b_n ),
\end{equation}
where o represents the output label and  \( x_{n} \) is the output of the last inner block. 

The trained BNN can be expressed in terms of cardinality constraints as explained in \citet{narodytska2018verifying}, by encoding each of the blocks of the BNN into Mixed Integer Linear Programming(MILP), refined into Integer Linear Programming(ILP) and further refined into Conjunctive Normal Form(CNF) encodings. 

The Binarised Linear Layer as defined in (1) represents a linear constraint and can be expressed as: 
\begin{equation}
    y_i = \langle a_i, x_k \rangle + b_i, \quad i = 1, \ldots, n_{k+1},
\end{equation}
where \( a_i \) is the \( i \)th row of the matrix \( A_k \).
Similarly the  batch normalization layer defined in (2) that takes the output of the linear layer as input also represents a linear constraint that for the \( k \)th  block and \( i \)th  element, can be expressed as:
\begin{equation}
    \sigma_{k,i} z_i = \alpha_{k,i} y_i - \alpha_{k,i} \mu_{k,i} + \sigma_{k,i} \gamma_{k,i}, \quad i = 1, \ldots, n_{k+1}.
\end{equation}

The Binarisation layer as defined in (3), that implements the sign function represents implication constraints, given as:
\begin{equation}
    z_i \geq 0 \Rightarrow v_i = 1, z_i < 0 \Rightarrow v_i = -1, \quad i = 1, \ldots, n_{k+1},
\end{equation}
where  \( x_{k+1} = (v_1, \ldots, v_{n_{k+1}}) \) is the output of the \( k \)th block.

The MILP encodings for the inner block of the BNN represented by Equations (5),(6), and (7) include both real and integer values. The real part in the equation (7) after substituting (5) and (6) in it is rounded off to get ILP encodings as:
\begin{equation}
\langle a_i, x_k \rangle \geq C_i \Rightarrow v_i = 1, \langle a_i, x_k \rangle < C_i \Rightarrow v_i = -1, \quad i = 1, \ldots, n_{k+1}
\end{equation}
where all the variables have integer values. 

In the output block of the BNN the linear layer is encoded into MILP as above followed by an encoding for argmax as defined in (4), using the ordering relation between the logits  \( l_i \) for \(i = 1, \ldots, c\),  from the previous layer by introducing a set of Boolean variables \( b_{ij} \)'s such that
\begin{equation}
l_i \geq l_j \Leftrightarrow b_{ij} = 1, \quad i, j = 1, \ldots, c,
\end{equation}
where  \(c\) is the number of the classes in the classification task. 
The encodings for the argmax in the output block of the BNN as defined in (9) can be represented in terms of the input variables and encoded into ILP constraints by rounding off the difference of the real bias values as:
\begin{equation}
\langle a_i - a_j, x_n \rangle \geq [b_j - b_i] \Leftrightarrow b_{ij} = 1, \quad i, j = 1, \ldots, c.
\end{equation}
The sum of the boolean variables \( b_{ij} \) for a given i and  \(j = 1, \ldots, c\) will be equal to c if and only if all of the above implication constraints are true, hence the output variable \( o \) represented in ILP constraints as:
\begin{equation}
\sum_{j=1}^{c} b_{ij} = c \Rightarrow o = i, \quad i,j = 1, \ldots, c.
\end{equation}
will provide the output label for the given input. 

The ILP encodings of each of the blocks can either be directly translated into SAT to get the CNF encodings or further refined using Sequential Counters. Finally the encoding for the entire BNN will be the conjunction of the encodings of the individual blocks of the BNN including all the inner and the output blocks. For Example, a BNN with a single inner block and an output block can be represented as 
\begin{equation}
BNN(X,H,Y) = B_1  \wedge B_O,
\end{equation}
where \(X\), \(H\)  and  \(Y\) are the sets of the input, hidden \& output variables respectively while \(B_1\) and \(B_O\) are the encodings for the only inner \& the output block respectively. 

The inversion process can now be performed on the encoded CNF formula by appropriately constraining the output variables. The clauses implementing constraints on the output label are conjoined to the encoded CNF formula as
\begin{equation}
B_1  \wedge B_O \wedge O(Y=y)
\end{equation}
 where \(O\) is the constraint on the output variables in which the desired output label is set to true and the resulting CNF formula is sampled for satisfying assignments using CMSGen. Eventually if the resulting CNF formula is satisfiable, the inputs for the set output label can be reconstructed from the satisfying assignments to the input variables.

\end{document}